\documentclass{article}
\usepackage{spconf,amsmath,graphicx}
\usepackage{multirow}
\usepackage{subfig}   

\title{Bio-Inspired Spiking Convolutional Neural Network using Layer-wise Sparse Coding and STDP Learning}
\name{Amirhossein Tavanaei, Anthony S. Maida}
\address{University of Louisiana at Lafayette\\
The Center for Advanced Computer Studies, Bio-inspired AI Lab\\
301 E. Lewis Street, Lafayette, LA 70503, USA \\
\textit{\{tavanaei,maida\}@louisiana.edu}}
\begin{document}
%
\maketitle
\begin{abstract}
Hierarchical feature discovery using non-spiking convolutional neural networks (CNNs) has attracted much recent interest in machine learning and computer vision. 
However, it is still not well understood how to create a biologically plausible network of brain-like, spiking neurons with multi-layer, unsupervised learning.
This paper explores a novel bio-inspired spiking CNN that is trained in a greedy, layer-wise fashion. 
The proposed network consists of a spiking convolutional-pooling layer followed by a feature discovery layer extracting independent visual features. 
Kernels for
the convolutional layer are trained using local learning.
The learning is implemented using
a sparse, spiking auto-encoder representing primary visual features. 
The feature discovery layer extracts independent features by probabilistic, leaky integrate-and-fire (LIF) neurons that are sparsely active in response to stimuli. The layer of the probabilistic, LIF neurons implicitly provides lateral inhibitions to extract sparse and independent features.

Experimental results show that the convolutional layer is stack-admissible, enabling it to support a multi-layer learning. 
The visual features obtained from the proposed probabilistic LIF neurons in the feature discovery layer are utilized for training a classifier. 
Classification results contribute to the independent and informative visual features extracted in a hierarchy of convolutional and feature discovery layers. 
The proposed model is evaluated on the MNIST digit dataset using clean and noisy images. 
The recognition performance for clean images is above 98\%. The performance loss for recognizing the noisy images is in the range 0.1\% to 8.5\%  depending on noise types and densities. 
This level of performance loss indicates that the network is robust to additive noise.
\end{abstract}
\section{Introduction}
Hierarchical neural architectures resemble the ventral visual cortex of the primate brain~\cite{kruger2013}. Although deep learning~\cite{schmidhuber2015,lecun2015,jo2015,goodfellow2016deep} and specifically convolutional neural networks (CNNs)~\cite{krizhevsky2012,sainath2013} have shown breakthrough performance in pattern recognition, the question of learning the hierarchical architecture of biologically plausible neural layers has not been solved.   
Additionally,
most CNNs are trained by backpropagation, which cannot be computed locally. 
This paper, as the extended version of our previous research~\cite{tavanaei2017}, addresses the challenge of equipping spiking CNNs with biologically plausible multi-layer learning.
In addition to improved bio-realism,
spiking CNNs have better prospects for extremely low-power implementations in appropriate hardwares~\cite{cao2014}.

One of the challenges in using spiking CNNs is that they are difficult to train.
This can be seen among researchers who attempt to use spiking networks on neuromorphic hardware.
To avoid the difficulties of directly training a spiking CNN,
these networks are usually trained as a conventional (non-spiking) CNN and then,
after training, are translated to a spiking network~\cite{cao2014,matsugu2002,diehl2015}.
For instance, Cao et al. (2014) developed a spiking CNN by converting an already trained, rate-based CNN to a spike-based implementation~\cite{cao2014}. 
Diehl et al. (2015) extended the conversion method introduced in~\cite{cao2014} to reduce the performance loss 
during the conversion using a weight adjustment approach~\cite{diehl2015}. 
Our interest is in developing spiking CNNs which directly use multi-layer learning.

A number of spiking CNNs \cite{masquelier2007,wysoski2008,Beyeler2013a,kheradpisheh2016bio} trained by spike-timing-dependent plasticity (STDP) \cite{dan2006} currently exist.
One of their limitations is that they utilize only one trainable layer of unsupervised learning. That is, they avoid multi-layer learning.
The network of
Masquelier and Thorpe (2007), which is possibly the earliest spiking CNN, has this property \cite{masquelier2007,masquelier2010}.
It consists of a convolutional/pooling layer followed by a feature discovery layer
and a classification layer.
Only the feature discovery layer uses unsupervised learning.
Wysoski et al. (2008) used a similar design which extracted initial features using difference of Gaussian (DoG) filters in different orientations~\cite{wysoski2008}. 
This network also had only one trainable layer for unsupervised learning. 
Furthermore, neither of these networks trained the earlier feature extraction layer, but instead
used handcrafted Gabor or DoG filters. Recent extensions of~\cite{masquelier2007} providing multi-layer STDP-based networks still utilize the handcrafted filters for primary visual feature extraction~\cite{mccarroll2015bio,kheradpisheh2016stdp}. The recent works of~\cite{Panda2016a} and~\cite{lee2016training} develop the multi-layer learning in deep spiking networks using backpropagation. However, the backpropagation algorithm is not biologically plausible. 
Our paper introduces a novel bio-inspired 
spiking CNN
in which both the convolutional (feature extraction) and feature discovery layers are trained locally using layer-wise, unsupervised learning. Learning the convolutional layer corresponds to learning the convolutional filters using a sparse, spike-based visual feature coding. The feature discovery layer consists of probabilistic LIF neurons equipped with a probabilistic local learning rule to extract independent features for classification.

Our first contribution is to train convolutional filters with learned detectors acquired by a biologically plausible, state-of-the-art, 
sparse coding model. 
The acquired detectors represent the model receptive fields whose shapes resemble those found in primate visual cortex (area V1). 
Sparse representations, where each input state is coded by a few active units, are a compromise between extremely localized representations and fully distributed representations, while being easy to analyze~\cite{foldiak1990}. 
Each unit represents one component or feature of the pattern. 
The construction of sparse representations that resemble V1 receptive fields 
has been achieved by different methods such as feedforward Hebbian units interconnected by lateral anti-Hebbian feedback synapses~\cite{foldiak1990}, minimizing reconstruction error combined with a sparsity regularizer~\cite{olshausen1996}, independent components analysis (ICA)~\cite{bell1997}, and minimizing the energy function of neural activity distribution~\cite{rehn2007}. 
In terms of sparse spike code formation, recent work has used spiking networks to study the acquisition of visual sparse code comparable to visual features found in cortical area V1 using inhibitory interneurons~\cite{king2013}, ICA~\cite{savin2010}, mirrored STDP~\cite{burbank2015}, and canceling the correlations between filters (matching pursuit, MP)~\cite{perrinet2004}. Zylberberg et al. (2011) introduced the sparse and independent local network (SAILnet) of spiking neurons to produce the sparse representations of visual features for natural image patches represented by 1356 units~\cite{zylberberg2011}. Specifically, the output units exhibited three types of receptive fields:
``small unoriented features, localized and oriented Gabor-like filters, and elongated edge detectors"~\cite[p.\ 5]{zylberberg2011}. Based on this assumption that these detectors can be used as convulutional filters, we use a modified architecture of the SAILnet to train orientation selective convolutional kernels.

Our second contribution is to learn a feature discovery layer consisting of probabilistic neurons. This layer is stacked on the convolutional-pooling layer. The probabilistic neurons implement a winners-take-all competition to extract independent features. This layer uses an unsupervised probabilistic STDP learning rule.
Previous studies provide evidence that Bayesian analysis of sensory stimuli occurs in the brain~\cite{rao2002,doya2007,mozer2008,kording2004}. In Bayesian inference, the hidden causes are inferred using prior knowledge and the likelihood of the observation to obtain a posterior probability. The plausible learning rules based on the joint distribution of perceptions and hidden causes have attracted much interest in many areas such as adapting the spontaneously spike sequences close to the empirical distribution of actual ones~\cite{rezende2011}, expectation maximization (EM) approximation in recurrent SNNs~\cite{brea2011}, probabilistic information representation~\cite{zemel2004}, Markov chain Monte Carlo sampling~\cite{buesing2011}, and probabilistic association between neurons generated by STDP~\cite{pecevski2016}. Nessler et al (2009 and 2013) showed that their STDP variation implements a stochastic online EM algorithm~\cite{nessler2009,nessler2013}. Later, this STDP rule was used in reservoir computing~\cite{klampfl2013}, spatio-temporal pattern recognition in a hidden Markov model (HMM)~\cite{kappel2014}, and Gaussian mixture model (GMM) training embedded in the HMM states~\cite{tavanaei2016b}. One drawback of Nessler's STDP rule is that its `excitatory' connections use negative weights. Tavanaei et al (2016) modified Nessler et al's STDP rule, based on positive excitatory connections, for visual feature acquisition in a spiking CNN~\cite{tavanaei2016}. 
Here, the feature discovery layer in the spiking CNN is implemented as a single layer network of novel probabilistic neurons equipped with the probabilistic STDP. Relative likelihood of the proposed STDP components (LTP and LTD) specifies a Bayesian computation occurring in the synaptic weight plasticity. 

Finally, the pipeline of convolutional and feature discovery layers is trained layer-wise in a spiking framework that
results in a spiking CNN extracting independent features for pattern recognition task. Our approach falls within the framework of a stacking paradigm for deep belief networks because
the filter weights for the feature extraction layer are frozen after they are trained.
The spiking CNN used in this study is a multi-layer architecture of visual feature extraction modules
that have two layers of bio-inspired learning.

\section{Method}
\label{sec:1}
This section specifies the details of our spiking CNN architecture.
As previously mentioned, the two learning layers
are trained in a greedy, layer-wise fashion. 
The first learning layer is trained using the sparse coding algorithm.
These trained weights are then used as filters in the convolutional layer of the spiking CNN\@.
The convolutional layer projects to a pooling layer to reduce the feature map size 
and to improve invariance to local translation and scale. 
Spike trains generated by the pooling layer neurons project to the feature discovery layer. 
This layer is trained 
according to the spikes received from the pooling layer using an unsupervised STDP rule.

Section~\ref{cnn1} describes how the convolutional filters are trained and Section~\ref{subsec:convolution} describes how they are used to transfer spike trains to the pooling layer (Section~\ref{cnn3}). Section~\ref{cnn4} explains the feature discovery layer's architecture and its learning rule to provide visual features for classification as described in Section~\ref{cnn5}.

\subsection{Convolutional Filters} 
\label{cnn1}
\begin{figure*}

\begin{minipage}[c]{0.5\textwidth}
\centering
\subfloat[]{
\includegraphics[scale=.7]{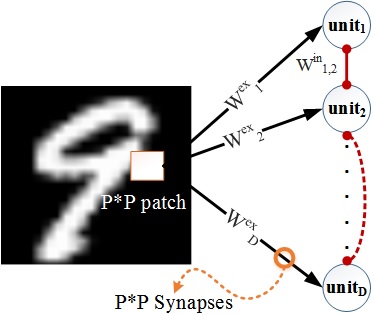}}

\end{minipage}
\begin{minipage}[c]{0.5\textwidth}
\centering
\subfloat[]{
\includegraphics[scale=0.8]{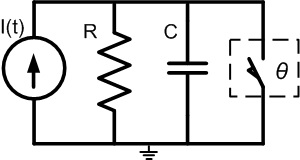}}

\end{minipage}
\caption{a: SAILnet architecture for training $D$ representation units to become detectors
for V1-like features.
The $D$ filters are represented as $D$ excitatory weight sets, $W^{\mathrm{ex}}$. 
Each representation unit receives $p^2$ pixel intensity values. b: RC circuit modeling an LIF neuron with injected current $I(t)$, resistor $R$, capacitor $C$, and a threshold-based ($\theta$) switch that instantly discharges the circuit to 
reset potential (zero) and emits a spike.
}
\label{fig:sailnet}
\end{figure*}

The convolutional layer is represented by $D$ filters.
It extracts visual features across the image,
resulting in one feature map from each filter. 
The SAILnet architecture was modified and used to train the required filters. 
It learns to represent an image as a set of
sparse, distributed features of visual area V1. 
The network has an input layer and a representation layer as shown in Fig.~\ref{fig:sailnet}a. The representation layer is fully connected laterally by inhibitory weights, denoted $W^{\mathrm{inh}}$. The LIF neurons in the representation layer receive an image patch with $p\times p$ pixels via fully connected excitatory weights ($W^{\mathrm{ex}}$). 
A set of $p^2$ excitatory weights projects to each neuron in the representation layer and determines a specific filter. 
Fig.~\ref{fig:sailnet}a shows the SAILnet architecture for training $D$ filters projecting an image patch to $D$ different spiking neurons. Each filter stands for synaptic weight sets connected to a unit. $D$ in our model is very smaller than the SAILnet.

\subsubsection{Stimulus presentation for SAILnet}  
A $p\times p$, normalized, non-spiking patch is presented to the network via fully connected excitatory synapses initialized 
with uniformly distributed noise in the range (0, 1). 
The stimulus is presented to the LIF neurons (simulated by circuit in Fig.~\ref{fig:sailnet}b) in the representation layer for $T=20$ simulation time steps ($T=20$ ms). 
The pixel intensity, $x_k$, in combination with lateral inhibition from the other neurons 
formulate the membrane potential of neuron $i$ as shown in Eq.~\ref{eq:sailnet}.

\begin{subequations}
\label{eq:sailnet}
\begin{align}
&\frac{dU_i(t)}{dt}+U_i(t)=I(t)\\
&I(t) = \sum_k W^{\mathrm{ex}}_{ik} x_k - \sum_{m\neq i} W^{\mathrm{inh}}_{im}z_m(t)
\end{align}
\end{subequations}


\noindent
For implementation,
the above continuous equation is discretized into 1 ms time steps.
The binary value $z_m(t)=1$ if and only if unit $m$ in the representation layer spikes at time $t$. 

\subsubsection{SAILnet learning rules}
The SAILnet algorithm incorporates excitatory weights ($W^{\mathrm{ex}}$), inhibitory weights ($W^{\mathrm{inh}}$), and thresholds ($\theta$) as adaptive parameters. $W^{\mathrm{inh}}$'s and $W^{\mathrm{ex}}$'s are initialized to zero and uniformly distributed noise in the range (0,1), respectively. 
Thresholds, $\theta_i$, are initialized to 5. The learning rules, which are applied after each 20 ms stimulus presentation, are given below.

\begin{subequations}
\label{eq:Zylberberg}
\begin{align}
\Delta W^{\mathrm{inh}}_{im} &= \alpha (n_i n_m - \rho^2)\label{eq:zyl1}\\
\Delta W^{\mathrm{ex}}_{ik} &= \beta n_i(x_k - n_i W^{\mathrm{ex}}_{ik})\label{eq:zyl2}\\
\Delta \theta_i &= \gamma (n_i - \rho)\label{eq:zyl3}
\end{align}
\end{subequations}


\noindent
$n_i$ (or $n_m$) is the number of spikes emitted during a stimulus presentation. 
Eq.~\ref{eq:zyl1} implements an anti-Hebbian rule among the inhibitory synapses. Eq.~\ref{eq:zyl2} implements a Hebbian rule among the feedforward synapses. 
The sparsity parameter, $\rho$, is the target average value for the number of spikes per image patch. 
$\rho$ is set to 0.05 in the experiments. 
Learning rules~\ref{eq:zyl1} -- \ref{eq:zyl3} are obtained by the Lagrange function (Eq.~\ref{eq:lag}) where the 
reconstruction error is minimized while the output units have a low, fixed, average firing rate, $\rho$, 
with minimal correlation between neurons~\cite{zylberberg2011}.

\begin{equation}
\label{eq:lag}
L=\sum_k \big (x_k-\sum_i n_i W^{\mathrm{ex}}_{ik} \big )^2 + \sum_i \zeta_i (n_i-\rho) + \sum_{i \neq m}\phi_{im} (n_i n_m-\rho^2)
\end{equation}   

\noindent
In the above, $\zeta_i=-\gamma_i$ and $\phi_{im}=-\alpha_{im}$ are unknown Lagrange multipliers.

\subsubsection{Representation Properties}
The sparsity parameter, $\rho$, and inhibitory connections, $W^{\mathrm{inh}}$, ensure that activity of the representation neurons remains sparse and independent after training. 
For instance, after training, an image patch with $p^2=5^2$ pixels is represented as $p^2\cdot \rho=1.25$ spikes of activity across the representation layer. 

\subsection{Convolution}
\label{subsec:convolution}
We now return to describing 
how the convolutional filters are used.
The convolutional layer uses weight sharing to reduce the number of parameters 
as in a conventional CNN\@.
However, the information in a spiking CNN is transferred via spike trains instead of real values. 
The input image is represented by spike trains whose rates are determined
by the normalized gray scale pixel intensities in the range (0, 1).
The image is presented to the network for $T=20$ ms divided into 1 ms time steps.

\begin{figure*}
\centering
\includegraphics[scale=.65]{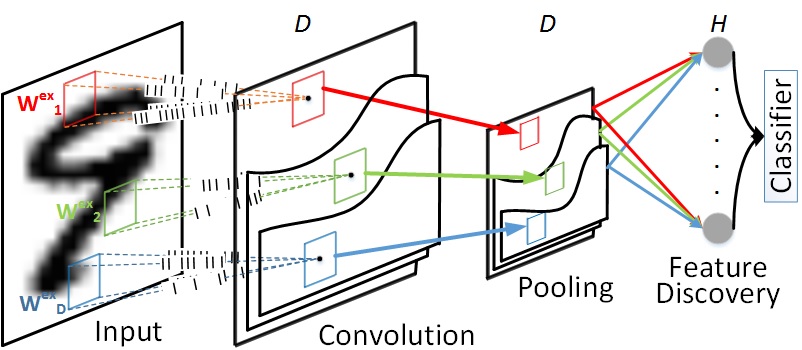}
\caption{Spiking CNN architecture. 
The first layer convolves the $D$ pretrained filters to the image 
(presynaptic spike trains) to generate $D$ feature maps. 
A patch in the input image is represented by a neuron in the feature map. 
The second layer is a non-adaptive, max-pooling layer. The third layer is the fully connected, feature discovery layer whose output is sent to a classifier.
}
\label{fig:convolution}
\end{figure*}

The first layer of the network (labeled convolution), seen in Fig.~\ref{fig:convolution}, shows the convolution process.
To generate feature maps, a set of $D$ filters are independently
convolved with the image over the 20 time step presentation interval (explained in Section~\ref{convneuron}). 
The values of the parameters, like $D$, are given in Table~\ref{tab:params}.
Once trained, these filter weights are frozen and no longer trained
when used in our spiking CNN\@.
When a filter is convolved with the input image, it generates its own feature map representing a specific image characteristic. 
Each image is represented by $D$ feature maps corresponding to the $D$ approximately independent filters.

\subsubsection{Convolutional neuron model}
\label{convneuron}
An LIF neuron in a feature map accumulates the convolution results for a particular image patch over $T=20$ time steps and fires a spike when reaching threshold. 
The membrane potential is reset to the resting potential (zero) upon firing. 
Eq.~\ref{eq:conv} calculates the membrane potential, $U$, of neuron $m$ in feature map $k$ at time $t$.
\begin{subequations}
\label{eq:conv}
\begin{align}
&\frac{dU_m^k(t)}{dt}+U_m^k(t)=I_m^k(t)\label{eq:conv1}\\
&\mathrm{if} \ \ U_m^k(t) \geq \theta^{\mathrm{conv}}, \mathrm{then} \ \ U_m^k(t)=0 \label{eq:conv2}
\end{align}
\end{subequations}
%

\noindent
$I_m^k$ is defined below.

\begin{equation}
I_m^k(t)=\sum_{i=1}^p \sum_{j=1}^p W^{\mathrm{ex}}_k(i,j)\cdot s_m(i,j,t)
\end{equation}

\noindent 
The $I_m^k(t)$ value determines the convolution of the filter, $W^{\mathrm{ex}}_k$, and the presynaptic spike train, $s_m$, in $T=20$ time steps. This value is considered as an injected current in the underlying RC circuit, which simulates the LIF neuron. 
$s_m(i,j)$ is the binary-valued spike train conveying a pixel value located in $(i,j)$ coordinate of patch $m$. 
The inputs to each neuron are represented as $p^2$ spike trains with $\lambda_{ij}$ rate parameters. At time step $t$, the expected value of the injected current, obtained by filter $k$ for neuron $m$, is given by
\begin{equation}
E[I_m^k(t)] = \sum_{i=1}^p \sum_{j=1}^p W^{\mathrm{ex}}_k(i,j)\cdot \lambda_{ij}
\end{equation}
This is analogous to $I=(\mathbf{w}^{\mathrm{ex}})^T\cdot \mathbf{x}$ ($\mathbf{w}^{\mathrm{ex}}$ is a filter and $\mathbf{x}$ denotes pixel intensities) in which the $\mathbf{x}$ is scaled to the average firing rates. Thus, the convolution over $T$ time steps is able to simulate the conventional convolution.

The last item in the convolutional layer is the stride value, $l_c$, which controls the sliding overlaps of a filter over the image and consequently the feature map size. A convolutional network with stride $l_c=1$ transfers the original image ($r\times c$ pixels) to feature maps with $r \times c$ neurons (by considering borders) or $(r-p+1)\times (c-p+1)$ neurons (by ignoring borders). Larger strides result in smaller feature maps. 

\subsection{Pooling}
\label{cnn3}
The pooling layer selects a neuron with the highest activity in a square neighborhood of $l_p\times l_p$ 
presynaptic neurons (max pooling). 
Parameter $l_p$ also serves as the stride value for the pooling layer. 
Activities of the neurons in a feature map can be characterized by their spike rate. 
Hence, a feature map with size of $r\times c$ is transferred to a smaller feature map with size of $r/l_p \times c/l_p$. 
The max pooling layer helps provide robustness against local changes in translation and scale. 
The pooling layer is shown as 
the second layer of the network (labeled pooling) in Fig.~\ref{fig:convolution}. 
The connections from the convolutional maps to the pooled maps are non-adaptive.

\subsection{Feature Discovery Layer}
\label{cnn4}
The spike trains emitted from the pooling layer carry diverse visual features of the image 
distributed across the $D$ feature maps. 
The feature discovery layer processes these features to recognize and discover more complex features. 
This layer is implemented in a fully connected feedforward fashion using $H$ output neurons. 
In our experiments, $H$ can be either 8, 16, 32, 64, 128, 256, or 512 (see Table~\ref{tab:params}).
Thus, the feature discovery layer reduces the dimension of the information in the feature maps. The third layer of the network in Fig.~\ref{fig:convolution} 
shows the feature discovery layer where the units receive spike trains emitted from the pooled feature maps.
 
The units in this layer are probabilistic LIF neurons such that their firing times depend on both membrane potential and firing probability. 
The next sections explain the probabilistic LIF neuron model, learning process, and stimulus representation of the network architecture.

\subsubsection{Feature discovery neuron model}
The units in this layer are LIF neurons (Eq.~\ref{eq:conv}). However, their firing times depend on both the usual threshold, $\theta^{\mathrm{h}}$, (Eq.~\ref{eq:conv2}) and firing probability in combination with a second threshold, $\theta^{\mathrm{p}}$.
A neuron fires if its membrane potential reaches the $\theta^{\mathrm{h}}$ and its  firing probability is greater than the $\theta^{\mathrm{p}}$.
The probability of neuron $h$ firing at time $t$ given input spike vector $\mathbf{y}_t$ 
and weights $W$ from the pooling layer to the hidden layer is given by Eq.~\ref{eq:posterior}.

\begin{equation}
\label{eq:posterior}
P(z_h(t)=1|W, \mathbf{y}_t) = \frac{e^{W_h^T \cdot \mathbf{y}_t}}{\sum_{j=1}^H e^{W_j^T \cdot \mathbf{y}_t}}
\end{equation}

\noindent
The above equation is a softmax function of the net input to the units at time $t$.
For units that reach threshold, $\theta^\mathrm{h}$, they fire according to this softmax probability,
although mutual exclusivity is not enforced. The softmax probability implements inhibition in this layer.
%
%
Each unit in the feature discovery layer has a chance to fire according to the probability given.

Furthermore, the probability in Eq.~\ref{eq:posterior} specifies that the firing probability of neuron $h$ is a conditional probability given the membrane potentials ($U(t)$) computed in all the neurons:
\begin{equation}
\label{eq:ind1}
P\{z_h(t)=1|U_1(t),U_2(t),...,U_H(t)\}
\end{equation}
Since the neurons in the feature discovery layer fire independently, the joint distribution of the case in which $J$ neurons fire at the same time, $t$, is calculated by
\begin{equation}
\label{ind2}
P\big (z_1(t)=1,...,z_J(t)=1|U_1(t),...,U_H(t)\big ) = 
\end{equation}
\begin{equation*}
\frac{e^{\sum_{j=1}^JU_j(t)}}{\big ( e^{U_1(t)}+...+e^{U_H(t)}\big )^J}
\end{equation*}
such that by increasing $J$ from 1 up to $H$, the probability value decreases. This follows from Eq.~\ref{eq:posterior} by multiplying the firing probabilities of individual neurons because they are conditionally independent. The features discovered by the neurons in this layer exhibit low correlation. In contrast, the LIF neurons, which only fire based on their membrane potential, can fire simultaneously whenever they reach the threshold. Because the neural firing can be simultaneous, there may be a high correlation between acquired features.

%

\subsubsection{Learning}
We use the STDP rule defined in Eq.~\ref{eq:stdp} for feature acquisition.
\begin{equation}
\label{eq:stdp}
\Delta w_{hi} = 
\begin{cases}
&a^+e^{-w_{hi}}, \ \ \ \mathrm{if} \ i \ \mathrm{fires \ in } \ [t-\epsilon , t]\\
&-a^-, \ \ \ \ \ \ \  \ \mathrm{otherwise.}
\end{cases}
\end{equation}
$a^+$ and $a^-$ are LTP and LTD learning rates such that $a^+/a^-=4/3$. 
When a postsynaptic neuron $h$ fires, an STDP event occurs. 
LTP occurs if the presynaptic neuron $i$ fires briefly (e.g. within $\epsilon=5$ ms) before the postsynaptic neuron $h$. 
Otherwise, LTD occurs. 
The current value of the synaptic weight in the LTP rule controls the changed value. 
It is proven in our previous study~\cite{tavanaei2016} that the synaptic weights are directly related to the log odds with respect to the probability at which LTP occurs (derivations are provided in~\cite{tavanaei2016}). Weights fall in the range (0, 1). 
The synaptic weight, $w_{hi}$, connecting presynaptic neuron $i$ and postsynaptic neuron $h$ is defined as

\begin{equation}
\label{prob}
w_{hi}=\mathrm{ln}\big (\frac{a^+}{a^-}\big )+\mathrm{ln} \big ( \frac{P(y_i=1|z_h=1)}{1-P(y_i=1|z_h=1)} \big )
\end{equation}

\noindent
where $P(y_i=1|z_h=1)$ is the LTP probability and $[1-P(y_i=1|z_h=1)]=P(y_i=0|z_h=1)$ is the LTD probability. Minimum and maximum LTP probabilities (for the synaptic weights 0 and 1) are controlled by the relative learning rate ($a^+/a^-$). If $a^+=a^-$, the LTP probability is shown as the logistic sigmoid function (Eq.~\ref{eq:sig}). The log odds value, in this case, specifies the minimum probability of 0.5 for LTP. The minimum LTP probability of 50\% shows that the synapses are able to undergo the LTP even if their weights are zero (currently inactive synapses).

\begin{equation}
\label{eq:sig}
P(y_i=1|z_h=1) = \frac{1}{1+e^{-w_{hi}}}
\end{equation}

Weights are initialized randomly between 0 and 1. After training, the weights are fixed to be utilized for complex feature discovery task.

\subsubsection{Bayesian inference}
The probability of LTP versus LTD determines a binary classification in which the LTP probability can be formulated by the Bayesian theorem~\cite{bishop2006} as follow:
\begin{equation}
\label{eq:bayes}
P(y_i=1|z_h=1) = 
\end{equation}
\begin{equation*}
\frac{P(z_h=1|y_i=1) P(y_i=1)}{P(z_h=1|y_i=1)P(y_i=1)+P(z_h=1|y_i=0)P(y_i=0)}
\end{equation*}
where, $P(y_i=1)$ can be interpreted as the presynaptic Poisson spike rate such that $P(y_i=0)=1-P(y_i=1)$. The firing probability, $P(z_h=1|\mathbf{y})$, defined in Eq.~\ref{eq:posterior}, depends exponentially on the neuron's membrane potential~\cite{jolivet2006,nessler2013}.
Additionally, Eq.~\ref{eq:bayes} can be written as
\begin{equation}
\label{bayessig}
P(y_i=1|z_h=1)=\frac{1}{1+e^{-q}}
\end{equation}
\begin{equation*}
\label{sigsig}
q = \ln \big (\frac{P(y_i=1|z_h=1)}{P(y_i=0|z_h=1)} \big ) = \ln \big (\frac{P(y_i=1|z_h=1)}{1-P(y_i=1|z_h=1)} \big )
\end{equation*}
which is equivalent to Eq.~\ref{eq:sig} obtained at equilibrium point of the probabilistic STDP rule. Therefore, the odds ratio of the LTP reflects Bayesian computation (Eq.~\ref{eq:bayes}) in which the likelihood of the LTP is obtained by a logistic sigmoid function of the LTP/LTD likelihood ratio.

\subsubsection{Input Representation in Feature Discovery Layer}
Each stimulus is represented by $H$ neurons. The feature discovery layer receives spike trains emitted from neurons in the pooling layer through trained weights, $W$. Thus, an image can be recoded to a vector with $H$ attributes. The vector attributes are measured as neural activities of the feature discovery layer. The neural activity can be specified as neuron's membrane potential or emitted spike train. In this paper, we use accumulated membrane potential that is obtained by integrating the net input of neuron over $T=20$ time steps.

\subsection{Classification}
\label{cnn5}
 The samples with $H$ attributes obtained by the feature discovery layer are used for training the classifier. In this investigation, we use support vector machine (SVM) classifiers to show the network ability in extracting independent and prominent features. The SVM classifiers allocate hyper-planes with maximum margins between class pairs of the data in a supervised learning fashion. 
 
\begin{figure*}
\centering
\includegraphics[scale=.7]{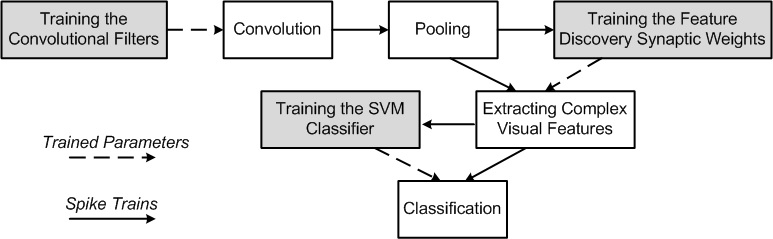}
\caption{The spiking CNN architecture for pattern recognition. The network is trained layer-wise (shown by shaded boxes).}
\label{fig:CSNNprocedure}
\end{figure*}

The procedural architecture of the spiking CNN is shown in Fig.~\ref{fig:CSNNprocedure}. The convolutional and feature discovery layers are trained layer-wise. The classifier is trained on samples obtained from the feature discovery layer.

\section{Experiments and Results}
The MNIST dataset containing normalized handwritten digits was used in the experiments. 
An MNIST image consists of $28\times 28$ ($r\times c$) grayscale pixel values. 
Each image was divided into overlapped $5\times 5$ ($p\times p$) patches with a stride of 1. 
For simplicity, we ignored image borders (margin size=$\lfloor p/2 \rfloor$). 
Thus, each image was represented by $(r-p+1)\times (c-p+1)=24^2=576$ overlapped patches. 

The training and information flow through the network are as follows.
First, the convolutional filters are trained to represent the V1 visual features. 
Second, the filters generate the feature maps by the convolutional and pooling layers. 
Third, the feature discovery layer is trained to extract complex features by combining the preliminary visual features. 
Finally, the SVM classifies the samples represented by the feature discovery layer. 
The parameters used in the experiments are shown in Table~\ref{tab:params}.

\begin{table}[]
\centering
\footnotesize
\caption{Parameter values of the spiking CNN used in the simulations.}
\label{tab:params}
\begin{tabular}{|l|l|l|}
\hline
\multicolumn{1}{|c|}{\textbf{Parameter}} & \multicolumn{1}{c|}{\textbf{Description}} & \multicolumn{1}{c|}{\textbf{Value}} \\ \hline
$D$                                        & \# of convolutional filters                       & 16, 32, or 64                          \\ \hline
$T$                                        & Simulation time                           & 20 ms                               \\ \hline
$p\times p$                                        & Patch size                                & $5\times5$                                   \\ \hline
$r$                                        & \# of image rows                                 & 28                                  \\ \hline
$c$                                        & \# of image columns                              & 28                                  \\ \hline
$\alpha$                                    & Hebbian learning rate                     & 0.01                                \\ \hline
$\beta$                                    & Anti-Hebbian rate                         & 0.0001                              \\ \hline
$\gamma$                                     & Threshold adjustment rate                 & 0.02                                \\ \hline
$\rho$                                      & Average firing rate                       & 0.05                                \\ \hline
$C$                                        & Capacitance                                 & 0.001 F                                   \\ \hline
$R$                                        & Resistance                                  & $1 \Omega$                                   \\ \hline
$l_c$                                       & Convolution stride                        & 1                                   \\ \hline
$l_p$                                       & Pooling stride                            & 2                                   \\ \hline
$\theta^{\mathrm{conv}}$               & Convolution threshold                     & 1                                   \\ \hline
$H$                                        & \# of feature discovery units                   & $8,16,32,...,512$                   \\ \hline
$a^+$                                       & LTP amplification rate                    & 0.001                               \\ \hline
$a^-$                                       & LTD amplification rate                    & 0.00075                             \\ \hline
$\theta^{\mathrm{h}}$                  & Feature discovery neuron's threshold      & 0.5                                 \\ \hline
$\theta^{\mathrm{p}}$                  & Feature discovery softmax threshold      & 0.5                                 \\ \hline
$\tau$                  & Decay time constant for LIF neurons      & 1 ms                                 \\ \hline
\end{tabular}
\end{table}

\subsection{Convolutional Filter}
\begin{figure}
\centering
\includegraphics[scale=.4]{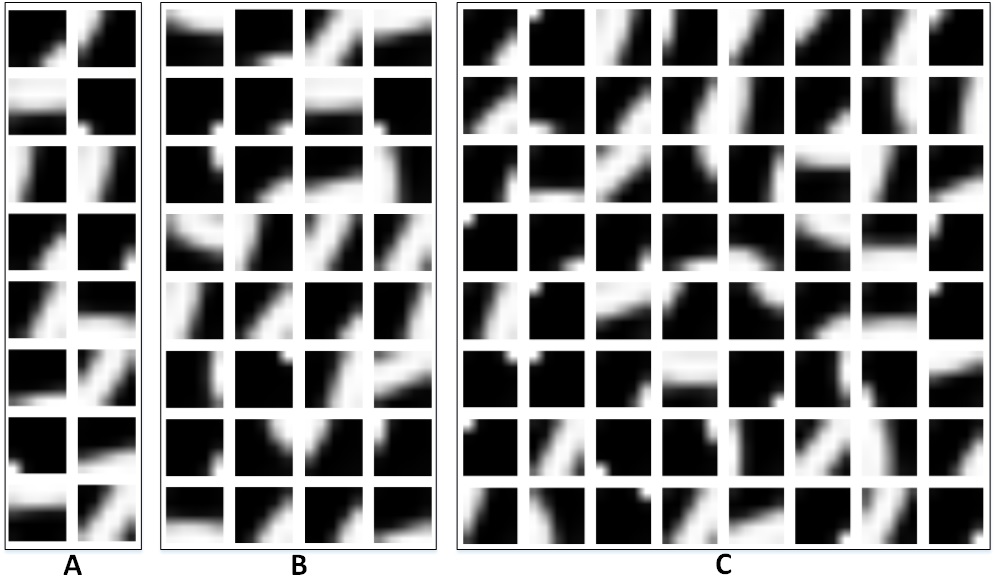}
\caption{Convolutional filters trained   
by the SAILnet algorithm to extract visual features.
Either (A) 16, (B) 32, or (C) 64 filters were trained. 
$5\times 5$ patches taken from MNIST images where used for the training.}
\label{fig:v1}
\end{figure}

The present experiment used 576 patches of size $5 \times 5$ sampled from the MNIST
data set.
The $28\times 28$ images were normalized to have zero mean and unit standard deviation.
3,000 images ($3,000\times 576 = 1,728,000$ patches) were used to train the convolutional filters. 
As seen in Fig.~\ref{fig:sailnet},
a patch (with 25 pixel values) was projected into $D$ representation neurons to train $D$ convolutional filters. 
Three filter sets of size $D=\{16, 32, 64\}$ were trained. 
One iteration is a batch of 3,000 image presentations. 10 training iterations were run and weights were updated after each patch representation (totally $17,280,000$ training trials). Fig.~\ref{fig:v1} shows trained filters (synaptic weights) discovering V1-like visual features. The filters are orientation selective. The filter sets containing more units depict a larger variety of orientations.      

The feedforward weights of the units ($5 \times 5$ squares) discover specific bars (light pixels) in different scales and orientations surrounded by the negative values (dark pixels). The positive and negative regions demonstrate the visual receptive fields in response to the image patches. The surrounding negative values discover edges corresponding to the filter orientation. 

\begin{figure}
\centering
\includegraphics[scale=.4]{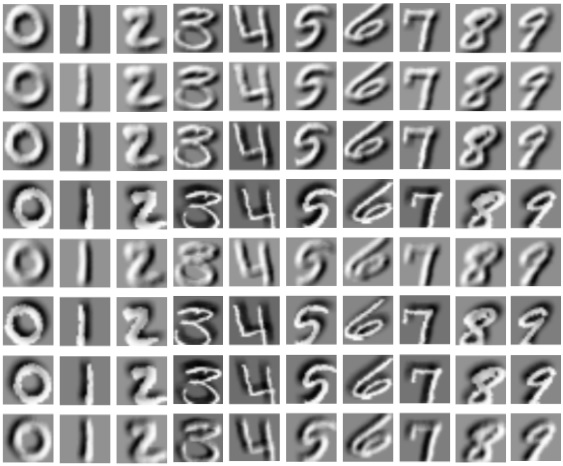}
\caption{Net inputs of eight randomly selected units in the SAILnet 
representation layer in response to the overlapped patches. Each column shows one digit in which eight convolutional filters (trained weights) were convolved to the image. Black, gray, and light pixels specify the negative, zero, and positive values, respectively.}
\label{fig:rf}
\end{figure}

Fig.~\ref{fig:rf} shows the net inputs
of eight randomly selected output units in response to the overlapped patches of the MNIST digits 0 to 9. The light pixels correspond to positive values showing the object. The gray pixels show zero value because the neuron does not receive any spike. The black pixels correspond to negative values next to the positive values in which edges were detected.  

\subsection{Convolution and Pooling Results}
The convolutional filters obtained above were used as the weights transferring presynaptic spikes to neurons in the feature maps. The presynaptic spikes are Poisson spike trains with rate of pixel intensity divided by 255 to obtain normalized intensities in the range (0, 1) for each of $T=20$ time steps. Each filter is convolved to the spike trains of the original image over 20 time steps and results in a feature map with $24\times 24 = 576$ postsynaptic LIF neurons. As the $D$ convolutional filters are independent, the feature maps are computed independently. For the case where 64 convolutional filters are applied, the convolutional layer represents an image by 64 independent feature maps such that each map contains 576 LIF spiking neurons. Spikes emitted from the neurons of a feature map indicate a particular visual feature. We applied 30,000 MNIST digits to the convolutional layer to generate representative spike trains for the next layer. For final evaluation, a complete version of the MNIST dataset containing 60K training and 10K testing samples was used.

\begin{figure*}
\centering
\includegraphics[scale=.8]{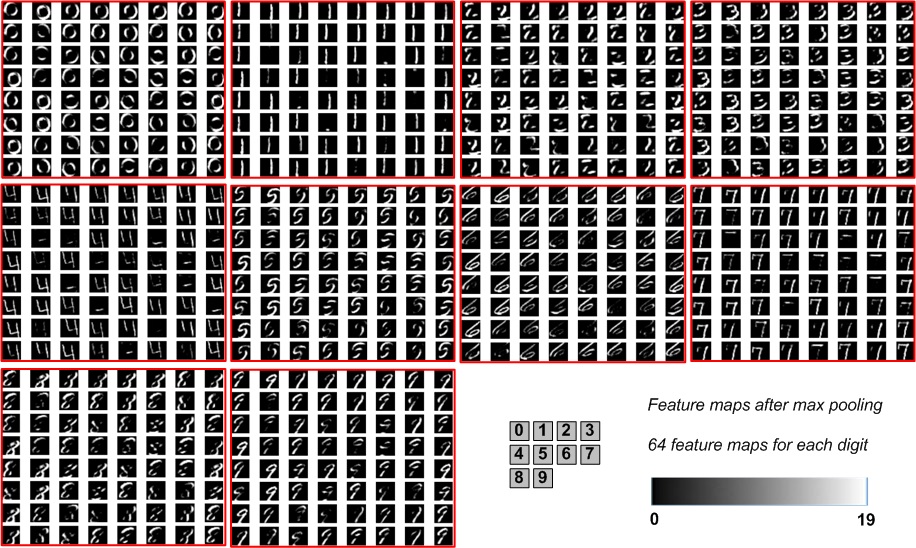}
\caption{64 feature maps of size $12\times 12$ for each of 10 digits (after max pooling). 
Each pixel represents the respective neuron's observed number of spikes
during the presentation interval. 
Each feature map indicates particular visual features. Digits 0 to 9 were randomly selected and applied to the  network.}
\label{fig:convRes}
\end{figure*}

The pooling layer receives spike trains from the non-overlapped neuron sets ($2\times 2$) in the feature maps and reduces the map size by using a max-pooling operation with stride of 2. Only one out of four neurons with maximum spike frequency is selected. Thus, a feature map with $24\times 24$ neurons is converted to a map with $12\times 12$ neurons. Thus, the cardinality of the feature set provided by the pooling layer is $D\times 144$.
In our experiments, $D$ can be 16, 32, or 64, so the output of the pooling layer will
have a minimum of 2,304 features. Neurons in this layer also indicate the visual features through their spiking patterns over 20 time steps. Fig.~\ref{fig:convRes} shows 64 pooled feature maps obtained from randomly selected digits 0 to 9. Pixel intensity codes a neuron's (one out of 144) spike rate. The maximum intensity (white) denotes 19 spikes and the minimum intensity (black) denotes 0 spikes emitted from the LIF neurons in the pooled feature maps. The feature maps for each digit, shown in Fig.~\ref{fig:convRes}, exhibit particular visual features. For instance, 64 feature maps of the digit `zero' show different orientations, scales, and local translations of the curved lines composing the image of `zero'. Digit `four' also is represented by different vertical and horizontal lines. 

These results indicate that the network can be extended by another layer which either combines the extracted features or discovers more features representing more details of the original image.
     
\subsection{Feature Discovery Results}
The last unsupervised learning process is performed by the feature discovery layer after receiving spike trains from the pooling layer. Each hidden unit has $D\times 12\times 12$ inputs.
Weights in this layer are feedforward, fully connected, and initialized randomly in the range (0, 1). 
The set of feature discovery neurons represents a feature vector for each digit sample.
For different experiments, 
we varied the number of feature discovery 
units with $H=\{8, 16, 32, 64, 128, 256, 512\}$. 
We used the accumulated membrane potential of the feature discovery neurons as feature values
for feature analysis and 
for later computations.

\begin{figure}
\centering
\includegraphics[scale=.25]{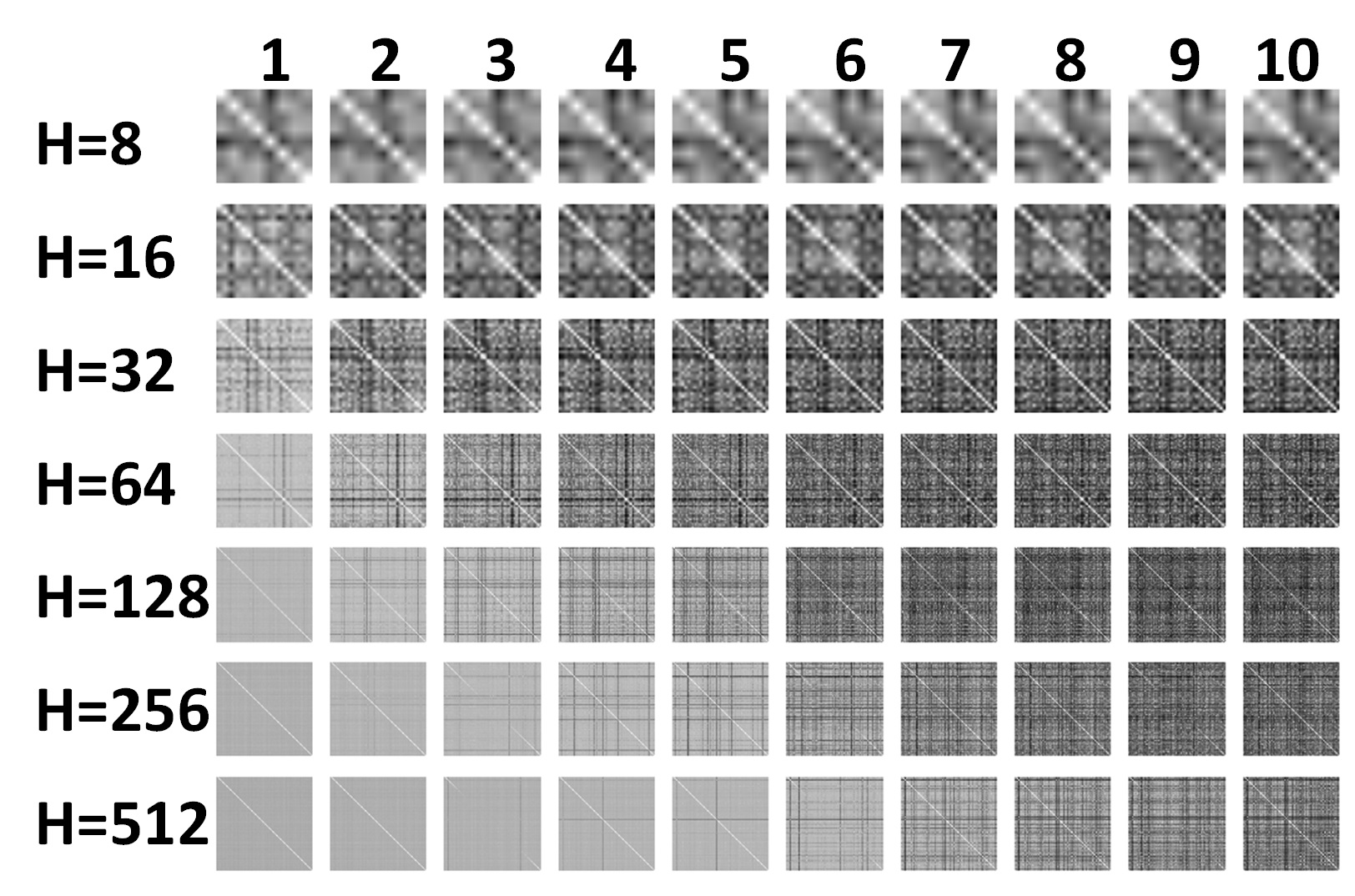}
\caption{Correlation matrices of the feature discovery weight sets for networks using $D=32$ feature maps and $H=\{8, 16, 32, 64, 128, 256, 512\}$ neurons over 10 iterations. A synaptic weight set stands for the synapses connecting neurons in the pooled maps to a postsynaptic neuron of the feature discovery layer. 
Neural activity becomes decorrelated with more training.}
\label{fig:corimage}
\end{figure}

Our main feature analysis result is that neural activity in the feature discovery
layer, as measured by membrane potentials, becomes decorrelated with training.
The weights, $W$, and spike trains emitted from the pooling layer provide input for the feature discovery neurons to generate feature values. Lower correlations between pairs of the weight sets, corresponding to the $H$ neurons, result in more independent feature values and consequently, better classification performance. Fig.~\ref{fig:corimage} shows correlation matrices ($H\times H$) computed for weight sets trained by spike trains emitted from neurons in $D=32$ pooled maps. The correlation values are shown in a symmetric matrix with elements between zero (black) and one (white). The correlation matrices show the progress of ten training iterations (3000 through 30,000 trials in increments of 3,000).
 Fig.~\ref{fig:corimage} shows that the neural activity of the feature discovery layer with $H=8$ or $16$ is highly correlated and does not change over training. However, the neural activity in the feature discovery layers with $H\geq 32$ units becomes decorrelated with more training.

\begin{figure}
\begin{minipage}[c]{0.55\textwidth}
\centering
\subfloat[]{\includegraphics[scale=.3]{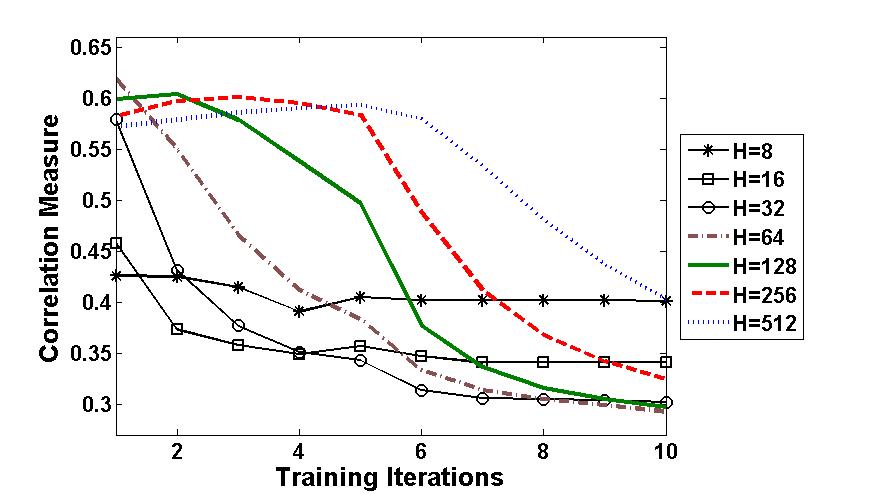}
}
\end{minipage}
\begin{minipage}[c]{0.5\textwidth}
\centering
\subfloat[]{\includegraphics[scale=.35]{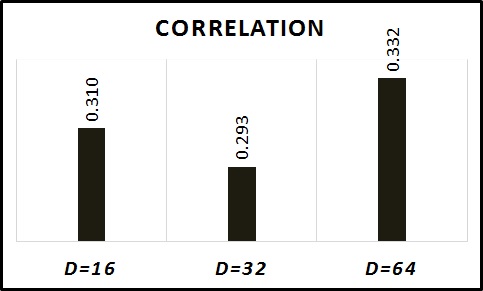}
}
\end{minipage}
\caption{a: Average correlations of synaptic weight sets versus training iterations for the network with $D=32$ feature maps. b: 
Lowest correlation of synaptic weight sets after training for the networks with $D=\{16, 32, 64\}$ feature maps.
}
\label{fig:corplot}
\end{figure}

Fig.~\ref{fig:corplot}a shows the average correlations between weight sets in ten training iterations for the networks with $D=32$ feature maps and $H=\{2^i, \ i=3...9\}$ feature discovery neurons. As described above, the average correlation decreases with training iterations and converges to an equilibrium. The correlation computed for the network with $H=8$ or $16$ neurons does not improve by training. The network with more neurons ($H=32, 64$ or $128$) has lowest average correlation. The plots in Fig.~\ref{fig:corplot}a show that the average correlations computed for the networks with $H=\{32, 64, 128, 256\}$ neurons are very close after 10 training iterations. Therefore, the network represents a stimulus in a feature space with minimum correlation between attributes. We also calculated the average correlations for the networks with $D=16$ and $64$ feature maps. As the trends in their graphs were virtually identical, we only showed the plots for the network with $H=32$ neurons. Fig.~\ref{fig:corplot}b shows the lowest average correlation calculated for the networks with $D=\{16, 32, 64\}$ feature maps. The network with 32 feature maps indicates minimum correlation between feature discovery neurons. 

%
%

\begin{figure}
\centering
A \includegraphics[scale=.5]{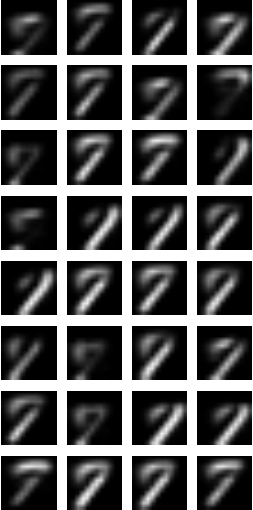} B \includegraphics[scale=.5]{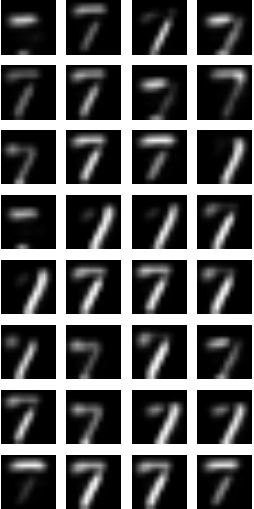} \\C \includegraphics[scale=.5]{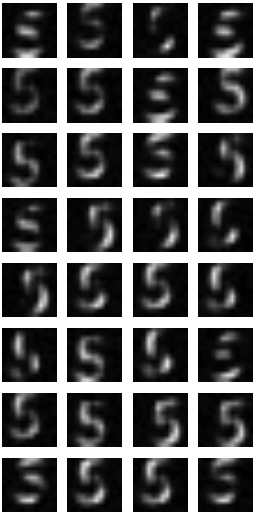} 
D \includegraphics[scale=.5]{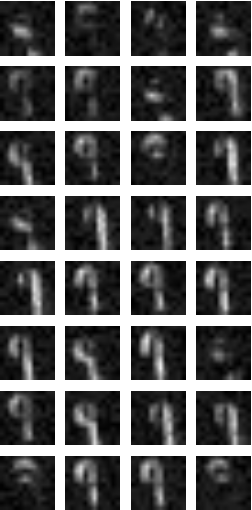} \\ E \includegraphics[scale=.5]{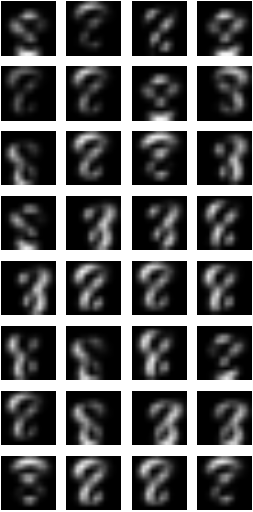}  \ F  \includegraphics[scale=.5]{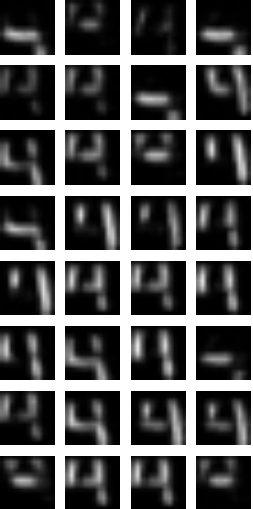}
\caption{Weights connecting pooled maps to six randomly selected neurons in the feature discovery layer. Each square shows the weights projecting from one pooled map (out of 32). Different digit variations appear in this layer. }
\label{fig:hiddenweights}
\end{figure}

Fig.~\ref{fig:hiddenweights} shows the trained synaptic weights connecting pooled maps (for $D=32$ convolutional filters) to six randomly selected output units. Each square in this image presents the synaptic weights connecting presynaptic neurons in one pooled map to one neuron. This layer represents different digit variations.

\subsection{Classification Results}
The visual features obtained from the feature discovery layer were applied to the SVM classifiers with four different kernel functions: linear, polynomial (degree=2 and 3) and radial basis function. Since all the SVMs showed close accuracy rates, we averaged over the SVM classifiers. Accuracy rates with standard errors were reported after each iteration using 5-fold cross validation. The initial results are reported for 30,000 MNIST digits. 

\begin{figure}
\centering
\includegraphics[scale=.35]{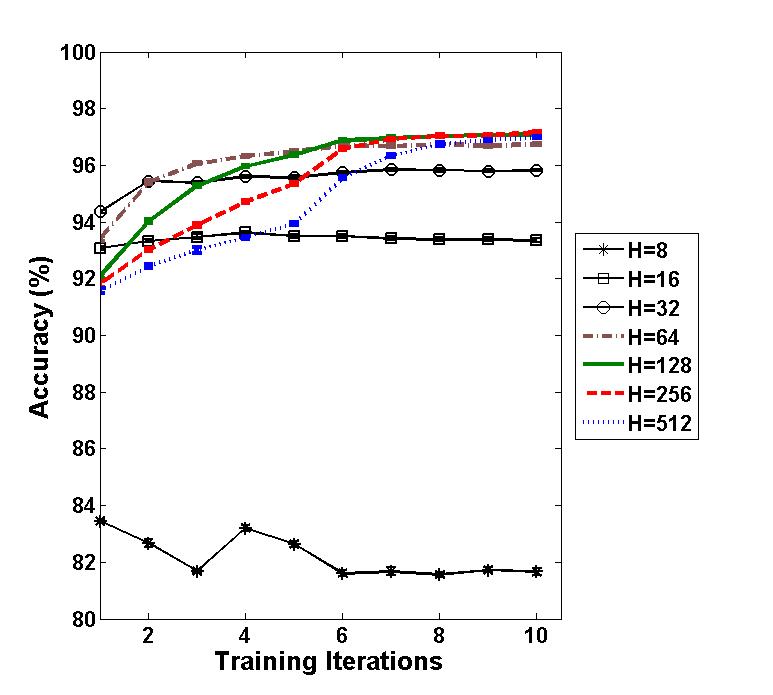}
\caption{Average classification accuracy for 10 training iterations. 
The convolutional layer has $D=32$ filters.}
\label{fig:d64acc}
\end{figure}

Table~\ref{tab:d64} shows the performance of the spiking CNN equipped with $D=\{16, 32, 64\}$ filters after training (10 iterations). The network with $H=8$ feature discovery neurons shows the lowest performance due to its small feature vector dimension. The networks with $H\geq 32$ show accuracy rates above 95\%. The overall accuracy rate of the network architecture with $D=32$ (above 97\%) is slightly better than the other models. Fig.~\ref{fig:d64acc} illustrates the model's performance at each training iteration for the network with $D=32$ feature maps. The weight adjustment progress regarding the handwritten digit classification suggests an analogous interpretation to the correlation reduction between postsynaptic neurons in the feature discovery layer (Fig.~\ref{fig:corplot}a). Additionally, the networks with $H=32$ and $64$ feature discovery neurons were trained fast but their performances did not improve after 2 and 6 training iterations, respectively. In contrast, the network with $H=512$ shows the lowest accuracy rate before 7 training iterations.  

\begin{table*}[]
\centering
\caption{The spiking CNN classification accuracy (with standard errors, $n=5$) obtained by averaging over accuracy rates reported for the SVM classifiers and 5-fold cross validation.}
\label{tab:d64}
\begin{tabular}{|c|l|l|l|l|l|l|l|l|}
\hline
\textbf{D}                   & \multicolumn{1}{c|}{\textbf{H}} & \multicolumn{1}{c|}{\textbf{8}} & \multicolumn{1}{c|}{\textbf{16}} & \multicolumn{1}{c|}{\textbf{32}} & \multicolumn{1}{c|}{\textbf{64}} & \multicolumn{1}{c|}{\textbf{128}} & \multicolumn{1}{c|}{\textbf{256}} & \multicolumn{1}{c|}{\textbf{512}} \\ \hline
\multirow{2}{*}{\textbf{16}} & \textit{Acc}                    & 82.25                           & 92.99                            & 95.69                            & 96.75                            & 96.97                             & 97.14                             & 96.95                             \\ \cline{2-9} 
                             & \textit{SE}                    & 0.25                            & 0.12                             & 0.12                             & 0.11                             & 0.08                              & 0.08                              & 0.06                              \\ \hline
\multirow{2}{*}{\textbf{32}} & \textit{Acc}                    & 81.66                           & 93.35                            & 95.82                            & 96.73                            & 97.09                             & 97.15                             & 96.97                             \\ \cline{2-9} 
                             & \textit{SE}                    & 0.26                            & 0.16                             & 0.10                             & 0.09                             & 0.07                              & 0.09                              & 0.07                              \\ \hline
\multirow{2}{*}{\textbf{64}} & \textit{Acc}                    & 82.96                           & 92.90                            & 95.37                            & 96.59                            & 96.89                             & 97.01                             & 96.95                             \\ \cline{2-9} 
                             & \textit{SE}                    & 0.16                            & 0.10                             & 0.14                             & 0.11                             & 0.09                              & 0.09                              & 0.08                              \\ \hline
\end{tabular}
\end{table*}

To assess the model's robustness to noise, we evaluated the network when classifying noisy MNIST digits. The network with $D=32$, $H=128$, and polynomial-2 SVM, which reported 97.5\% accuracy for clean images, was evaluated. 3,000 images that were not used for training were distorted by three levels of Gaussian noise with 4, 9, and 25 percent noise variances and three salt-and-pepper noises with 4, 9, and 25 percent noise densities. Table~\ref{tab:noisy} shows the accuracy rates of the network for noisy images in comparison with the case that the clean images were classified. The Gaussian noise reduced the performance by at most 1.2\% and salt-pepper noise showed a maximum 8.5\% performance loss. These results show that the network is fairly robust against additive noise.

\begin{table*}[]
\centering
\caption{Classification performance for noisy MNIST digits. Gaussian (Gauss) and salt-and-pepper (S-P) noises were added to the test image dataset. The network consists of $D=32$ feature maps (filters) and $H=128$ feature discovery units.}
\label{tab:noisy}
\begin{tabular}{|l|l|l|l|l|l|l|l|}
\hline
\multicolumn{1}{|c|}{\textbf{Noise}} & \multicolumn{1}{c|}{\textbf{\begin{tabular}[c]{@{}c@{}}No\\ Noise\end{tabular}}} & \multicolumn{1}{c|}{\textbf{\begin{tabular}[c]{@{}c@{}}Gauss\\ 4\%\end{tabular}}} & \multicolumn{1}{c|}{\textbf{\begin{tabular}[c]{@{}c@{}}Gauss\\ 9\%\end{tabular}}} & \multicolumn{1}{c|}{\textbf{\begin{tabular}[c]{@{}c@{}}Gauss\\ 25\%\end{tabular}}} & \multicolumn{1}{c|}{\textbf{\begin{tabular}[c]{@{}c@{}}S-P\\ 4\%\end{tabular}}} & \multicolumn{1}{c|}{\textbf{\begin{tabular}[c]{@{}c@{}}S-P\\ 9\%\end{tabular}}} & \multicolumn{1}{c|}{\textbf{\begin{tabular}[c]{@{}c@{}}S-P\\ 25\%\end{tabular}}} \\ \hline
\textbf{Accuracy (\%)}               & 97.5                                                                             & 97.4                                                                              & 97.2                                                                              & 96.3                                                                               & 97.3                                                                            & 96.5                                                                            & 89.0                                                                             \\ \hline
\end{tabular}
\end{table*}

\subsection{Control Experiments}
Control experiments were performed to further analyze the function of the feature discovery layer.
Specifically, we tried to assess the roles of the probabilistic STDP rule and 
the probabilistic LIF neuron, which implements the inhibition by a winners-take-all version of softmax, in explaining the classification accuracy results.
The architecture with $D=32$ filters and $H=128$ feature discovery units was selected for
further analysis. 

\begin{figure}
\centering
\includegraphics[scale=.35]{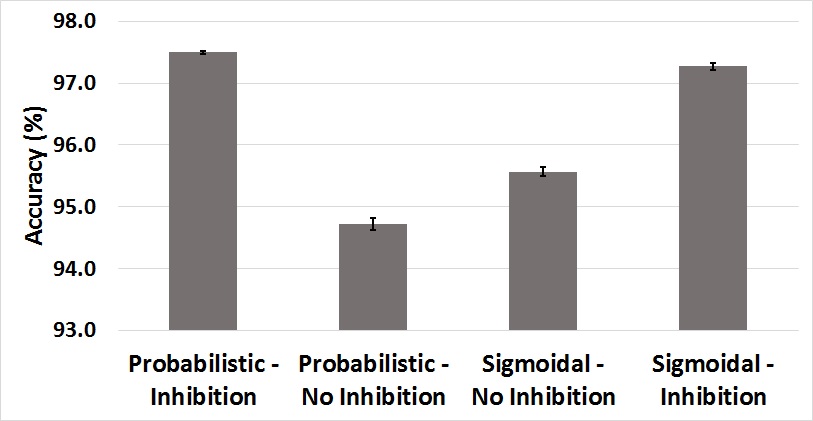}
\caption{Accuracy rates for the spiking CNN with 32 convolutional filters and 128 LIF/probabilistic neurons (No Inhibition/Inhibition) in the feature discovery layer equipped with the probabilistic/sigmoidal STDP rules recognizing clean MNIST digits.}
\label{fig:controlclean}
\end{figure}

\begin{figure}
\centering
\includegraphics[scale=.38]{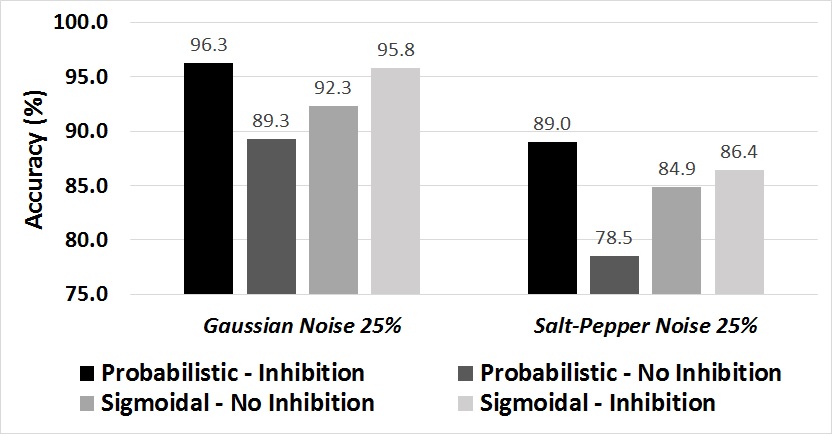}
\caption{Accuracy rates for the spiking CNN with 32 convolutional filters and 128 LIF/probabilistic neurons in the feature discovery layer equipped with the probabilistic/sigmoidal STDP rules recognizing noisy MNIST digits. Two types of Gaussian and salt-and-pepper noises were added to the MNIST digits.}
\label{fig:controlnoisy}
\end{figure}

This section provides more experiments to compare our feature discovery layer which is equipped with the probabilistic STDP rule and the probabilistic neuron model with the LIF neurons and the sigmoidal STDP used in the feature discovery layer of the network introduced by~\cite{masquelier2007}.
Four different feature discovery layers including 1) the proposed model, 2) LIF neurons equipped with the probabilistic STDP, 3) LIF neurons equipped with the sigmoidal STDP, and 4) probabilistic neurons equipped with the sigmoidal STDP were trained by the same training set used for the previous experiments. The performances of these spiking CNNs are shown in Figs.~\ref{fig:controlclean} and \ref{fig:controlnoisy} for clean and noisy images respectively. The network with the feature discovery layer consisting of the probabilistic neuron models equipped with the probabilistic STDP outperforms the other networks. As expected, the independent feature vectors represented by probabilistic neurons and ability of the firing probabilities (measured by the neurons) in feature enhancement (noise reduction) resulted in a significantly better performance of our spiking CNN. Additionally, the probabilistic neuron model performs better in the networks that utilize either the probabilistic or the sigmoidal STDP rules. That means, the inhibition in the feature discovery layer significantly (two-tailed T-test, $p^*<10e-4$) improves learning. The probabilistic STDP offers slightly better performance than the sigmoidal STDP where inhibition occurs. However, the sigmoidal STDP performs better for the LIF neurons (no inhibition). 

\section{Discussion}
The important part of multi-layer learning in a deep CNN occurs in the feature extraction hierarchy, where increasingly complex, discriminative, and invariant features are acquired~\cite{bengio2009,lecun2012}. Previous studies regarding the similarity between deep architectures and the feedforward structure of the ventral visual pathway of the primate brain ask about the biologically plausibility of the machine vision approach~\cite{kruger2013}. As biological neurons use sparse, stochastic, spike-based communication, a spiking network can be an appropriate starting point for modeling brain functionality.

\begin{figure*}
\centering
\includegraphics[scale=.79]{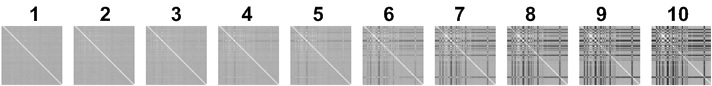}
\caption{Correlation matrices for the synaptic weight sets transferring spike trains to the LIF neurons without inhibition over 10 training iterations. The average correlation value after 10 iterations is 0.481.}
\label{fig:LIFcor}
\end{figure*}

\begin{figure*}
\centering
\includegraphics[viewport=0.4in 4.9in 8.5in 6.5in,clip=true,scale=.58]{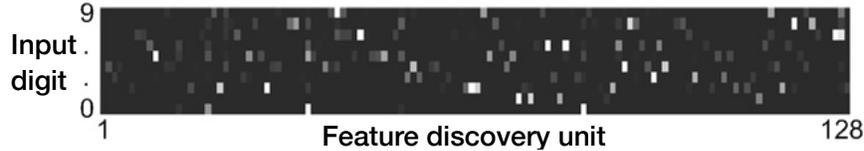}
\caption{Sparsity of average neural activity of $H=128$ 
feature discovery units. 
Lighter points indicate neuronal activity. 3000 images of the digits 0 -- 9 were used after training.
}
\label{fig:sparsity}
\end{figure*}

Our work seeks to study multi-layer unsupervised learning in the emerging field of spiking CNNs. We used systematic experiments to show the stackability properties of spike-based learning algorithms. Our spiking CNN resembles deep belief networks in the way that it is trained in a greedy, layer-wise fashion~\cite{hinton2006}. Once a layer is trained, its weights are frozen and its output spike trains serve as inputs to the next layer. For stacking to work, the learned representations must be admissible such that the acquired representations in a given layer are rich enough in content and statistical structure to support the next layer's learning. The proposed network consists of a convolutional layer, pooling operation, and a feature discovery layer. The convolutional layer, after training, showed promising intermediate results (Fig.~\ref{fig:convRes}) such that it exhibits the stack-admissibility property. Therefore, another layer can be stacked on this layer. The next layer was fully connected and trained with the probabilistic STDP rule. The average correlations demonstrated in Fig.~\ref{fig:corplot} specifies the model's ability to discover independent features. Furthermore, the feature discovery layer with many neurons converges to its minimum average correlation more slowly than the networks with fewer neurons. Additionally, the probabilistic neurons in this layer implemented inhibition by a winners-take-all competition governed by softmax. The inhibition controls the correlation between feature values obtained from the postsynaptic neurons. In contrast, the LIF neurons without inhibition do not control the correlation between features properly. The correlation matrices of the synaptic weight sets for the network with $D=32$ and $H=128$ LIF neurons in ten training iterations are shown in Fig.~\ref{fig:LIFcor}. This figure in comparison with the correlation matrices in Fig.~\ref{fig:corimage} (fifth row) exhibits higher feature dependency. Also, the average correlation value of the LIF neurons was 0.481 which is higher than the value reported for the probabilistic neurons (0.293). The independent features can also be interpreted as sparse and independent neural activities in response to specific stimuli. Fig.~\ref{fig:sparsity} shows the average neural activities of the feature discovery neurons ($H=128$) in response to 3000 images of the digits 0 through 9. Each digit activates only a subset of the postsynaptic neurons in the feature discovery layer. Therefore, dependency between the extracted features reduces while the inputs are sparsely coded.  

The classification performance showed that networks with $D=\{16, 32,64\}$ filters and $H=\{64, 128, 256,512\}$ feature discovery units produced high accuracy rates (average 97\%). The network with few trainable parameters was trained quickly. For example, the network with $D=16$ showed around 93.5\% accuracy while the network with $D=64$ showed 92\% average accuracy at the first training iteration. The network with few feature discovery units ($H=64$) was trained faster but its learning capacity was limited and its performance plateaued after seven iterations. Thus, the networks should have enough filters to extract a sufficiently complete feature set. Also, the number of feature discovery units control the network capacity for learning the complex features.
 
The control experiments for the feature discovery layer showed that the probabilistic neuron model implementing inhibition significantly improves learning. The probabilistic STDP performs slightly better than the sigmoidal STDP when the inhibition was used. This improvement also has been confirmed by a convolutional network of IF neurons equipped with a heuristic form of inhibition in our previous study~\cite{tavanaei2016}. Finally, the feature discovery layer of the probabilistic neurons equipped with the probabilistic STDP outperformed the other networks (Figs.~\ref{fig:controlclean} and \ref{fig:controlnoisy}). 

The spike trains emitted by the pooling layer indicate sparse, visual features of the original image and the convolution operation can perform as a mean filter enhancing (denoising) the noisy image. Thus, the network can be robust to additive noise. Additionally, the feature discovery layer combines sparse and independent features through sparse, independent weight sets which are trained by the probabilistic STDP. The probabilistic LIF neurons equipped with the probabilistic STDP fire based on both their membrane potential and firing probability. Additive noise changes the neuron membrane potential but the softmax, that is implemented by the firing probability, normalizes it according to the overall background level. Thus, the firing probability controls the additive noise by treating it as a background signal. The final results regarding the model's performance in classifying noisy images showed that our network is robust to additive noise.     

Finally, the best performing network architecture ($D=32$, $H=128$, and polynomial-2 SVM) was trained and evaluated 
on the full version of the MNIST dataset~\cite{lecun1998} 
(60K training, 10K testing, $28\times 28$-pixel images). 
Our spiking CNN achieved 98.36\% accuracy. This performance compares favorably to recent work regarding MNIST digit recognition using spiking CNNs and spiking deep belief networks (DBNs). Diehl et al (2015) reported 99.1\% accuracy for their network converting an off-line trained CNN onto a spiking network~\cite{diehl2015}. One drawback of their work is that the spiking CNN is not trained directly. O'Connor et al. (2013) showed that a DBN of LIF neurons performs well (94.1\%)~\cite{o2013}. They also mapped an off-line DBN onto an event-driven SNN. Neftci et al. (2014) showed that a spiking restricted Boltzmann machine can be trained locally using an STDP learning rule~\cite{neftci2014}. They reported performance of 93.6\% on MNIST using an STDP approximation to the contrastive divergence algorithm. Our spiking CNN was trained via a  layer-wise, unsupervised strategy. The final features discovered by our network were classified with about 98\% accuracy.       

\section{Conclusion}
Our work presents one of the few successful spiking networks directly trained by multi-layer, unsupervised learning. The proposed bio-inspired spiking CNN trained by a local learning rule is a starting point for implementing brain-like computation.
Due to the need for a biologically plausible CNN, we proposed a spiking CNN. The multi-layer learning implemented in the spiking CNN includes two components: 1) a convolutional-pooling layer and 2) a feature discovery layer. The network was trained layer-wise. After training the first layer, the weights were frozen to be used for the next layer's immediate receptive fields. The results acquired from the convolutional layer showed sparse, independent visual features (resembling those found in V1) such that the stack-admissibility property of the layer was supported. The feature discovery layer discovered complex, independent features using probabilistic LIF neurons equipped with the probabilistic STDP rule. The high classification performance achieved from the initial experiments on clean and noisy MNIST data sets (maximum 97.5\% for clean images and minimum 89\% for noisy images) showed that the spiking CNN can extract effective and independent visual features while being robust to additive noise. The final experiment on the MNIST dataset (60K training and 10K testing sets) confirmed the initial success by accuracy rate of 98.36\%.  

The promising results obtained from the convolutional layer (V1 visual features) and feature discovery layer (independent features) warrant further study. Our future work includes 1) adding more convolutional-pooling layers stacked on the first layer and 2) implementing a final classifier using a spike-based, supervised learning rule following the hierarchical feature extraction components. The combination of STDP and anti-STDP in our previous study~\cite{tavanaei2017neurocomputing} can be a starting point for developing the supervised classifier. 

\bibliographystyle{IEEEbib}
\bibliography{amir}

\end{document}